\documentclass[acmsmall,screen]{acmart}
\AtBeginDocument{%
  \providecommand\BibTeX{{%
    \normalfont B\kern-0.5em{\scshape i\kern-0.25em b}\kern-0.8em\TeX}}}

\setcopyright{acmcopyright}
\copyrightyear{2023}
\acmYear{2023}
\acmDOI{}

\acmJournal{TOMM}
\acmVolume{}
\acmNumber{}
\acmArticle{}
\acmMonth{3}

\usepackage{times}
\usepackage{epsfig}
\usepackage{graphicx}
\usepackage{amsmath}

\usepackage{amssymb}
\usepackage{url}
\usepackage{graphicx}
\usepackage[font=small,labelfont={bf}]{caption}
\usepackage{placeins}
\usepackage[aboveskip=0pt,font=normalsize]{subcaption}
\usepackage{xcolor}
\usepackage{enumitem}
\usepackage{float}
\usepackage{multirow}
\usepackage{makecell}
\usepackage{booktabs}
\usepackage[export]{adjustbox}
\newcommand\crule[3][black]{\textcolor{#1}{\rule{#2}{#3}}}

\def\etal{\textit{et al.}}
\def\ie{\textit{i.e.}}

\usepackage{amsmath,bm}
\usepackage{amsmath,xcolor}
\usepackage{array}
\makeatletter
\renewcommand*\env@matrix[1][*\c@MaxMatrixCols c]{%
	\hskip -\arraycolsep
	\let\@ifnextchar\new@ifnextchar
	\array{#1}}
\makeatother

\begin{document}

\title{A Geometrical Approach to Evaluate the Adversarial Robustness of Deep Neural Networks}

\author{Yang Wang}
\affiliation{
	\institution{Dalian University of Technology}
	\city{Dalian}
	\country{China}}
\email{yangwang06@mail.dlut.edu.cn}

\author{Bo Dong}
\affiliation{
	\institution{Princeton University}
	\city{New Jersey}
	\country{USA}}
\email{bo.dong@princeton.edu}

\author{Ke Xu}
\affiliation{
	\institution{City University of Hong Kong}
	\city{Hong Kong}
	\country{China}}
\email{kkangwing@gmail.com}

\author{Haiyin Piao}
\affiliation{
	\institution{Northwestern Polytechnical University}
	\city{Shannxi}
	\country{China}}
\email{haiyinpiao@mail.nwpu.edu.cn}

\author{Yufei Ding}
\affiliation{
	\institution{University of California, Santa Barbara}
	\city{California}
	\country{USA}}
\email{yufeiding@cs.ucsb.edu}

\author{Baocai Yin}
\affiliation{
	\institution{Dalian University of Technology}
	\city{Dalian}
	\country{China}}
\email{ybc@dlut.edu.cn}

\author{Xin Yang}
\authornote{Corresponding authors.}
\affiliation{
	\institution{Dalian University of Technology}
	\city{Dalian}
	\country{China}}
\email{xinyang@dlut.edu.cn}

\renewcommand{\shortauthors}{Wang, et al.}

\begin{abstract}
Deep Neural Networks (DNNs) are widely used for computer vision tasks. However, it has been shown that deep models are vulnerable to adversarial attacks, \ie, their performances drop when imperceptible perturbations are made to the original inputs, which may further degrade the following visual tasks or introduce new problems such as data and privacy security. Hence, metrics for evaluating the robustness of deep models against adversarial attacks are desired. However, previous metrics are mainly proposed for evaluating the adversarial robustness of shallow networks on the small-scale datasets. Although the Cross Lipschitz Extreme Value for nEtwork Robustness (CLEVER) metric has been proposed for large-scale datasets (\textit{e.g.}, the ImageNet dataset), it is computationally expensive and its performance relies on a tractable number of samples. In this paper, we propose the Adversarial Converging Time Score (ACTS), an attack-dependent metric that quantifies the adversarial robustness of a DNN on a specific input. Our key observation is that local neighborhoods on a DNN's output surface would have different shapes given different inputs. Hence, given different inputs, it requires different time for converging to an adversarial sample. Based on this geometry meaning, ACTS measures the converging time as an adversarial robustness metric. We validate the effectiveness and generalization of the proposed ACTS metric against different adversarial attacks on the large-scale ImageNet dataset using state-of-the-art deep networks. Extensive experiments show that our ACTS metric is an efficient and effective adversarial metric over the previous CLEVER metric.
\end{abstract}



\maketitle

\section{Introduction}
In recent years, deep learning (DL) has widely impacted computer vision tasks, such as object detection, visual tracking and image editing. Despite their outstanding performances, recent studies~\cite{athalye2018obfuscated,Chen2017EAD,moosavi-dezfooli2018robustness,szegedy2013intriguing,athalye2018synthesizing,zhang2021object,Li2021,ding2022biologically} have shown that deep methods can be easily cheated by the adversarial inputs: inputs with human imperceptible perturbations to force an algorithm to produce adversary-selected outputs. The vulnerability of deep models to adversarial inputs is getting significant attention as they are used in various security and human safety applications. Hence, a robust adversarial performance evaluation method is needed for existing deep learning models. The $l_p$ norm-ball theory may be used to indicate the adversarial robustness of neural networks. Specifically, this theory suggests that there should exist a perturbation radius $l_p$-distortion $\Delta_p$ = $\|\delta{x}\|_p$~\cite{weng2018evaluating}, where any sample point $x$ within this radius would be correctly classified as true samples, and others would be regarded as adversarial ones. In other words, the smallest radius $\Delta_p$ (\textit{i.e.}, minimum adversarial perturbation $\Delta_{p,min}$) can be used as a metric to evaluate the robustness: a model with larger radius indicates that it is more robust. However, determining the $\Delta_{p,min}$ has been proven in~\cite{katz2017reluplex,2017Certifying} as an NP-complete problem. Existing methods mainly focused on estimating the lower and upper bounds of $\Delta_{p,min}$. While estimating the upper bound~\cite{goodfellow2014explaining,kurakin2016adversarial,carlini2017towards} is typically attack-dependent, easy-to-implement and computational lightweight, it often suffers poor generalization and accuracy. On the contrary, estimating the lower bound~\cite{Weng2018Towards,NIPS2018_7742} can be attack-independent but computational heavy. Moreover, the lower bound estimation often provides little clues for interpreting the prevalence of adversarial examples \cite{gehr2018ai2,engstrom2017a,papernot2017practical,NIPS2018_7742}. 

To address the above limitations, this paper presents a novel instance-specific adversarial robustness metric, the Adversarial Converging Time Score (ACTS). Unlike CLEVER~\cite{weng2018evaluating}, ACTS does not use an exact lower bound of minimum adversarial perturbation as a robustness metric. Instead, ACTS estimates the desired robustness based on the $\Delta_{p,min}$ in the direction guided by an adversarial attack. ACTS is resilient, which means if an attack method can deliver a $\Delta_{p,min}$ attack, then the estimated robustness by ACTS reflects the fact. The insight behind the proposed ACTS is the geometrical characteristics of a DNN-based classifier's output manifold. Specifically, given a $M$-dimensional input, each output element can be regarded as a point on a $M+1$ dimensional hypersurface. Adding adversarial perturbations can be regarded as forcing the original output elements to move to new positions on those hypersurfaces. The movement driven by effective perturbations should push all output elements to a converging curve (\textit{i.e.}, the intersection of two or more hypersurfaces), where a clean input is converted to an adversarial one. Since the local areas around different points on hypersurfaces have different curvatures, different clean samples require different time to be converged to adversarial examples. The proposed ACTS measures the converging time and use the time as the adversarial robustness metric. To summarize, this paper has the following contributions. We propose a novel Adversarial Converging Time Score (ACTS) method for measuring the adversarial robustness of deep neural networks. Our method leverages the geometry characteristics of a DNN's output manifolds, so it is effective, efficient and easy to understand. We provide mathematical analysis to justify the correctness of the proposed ACTS and extensive experiments to demonstrate its superiority under different adversarial attacks.

This paper is organized as follows. We first review the related work in Section~\ref{sec:related}. In Section~\ref{sec:method}, we describe the proposed method. Results from comparative experiments for different architectures and adversarial attack approaches are then given in Section~\ref{sec:exper}. 
And we make the conclusions and envision the future work in Section~\ref{sec:conclusion}.

\vspace{-4pt}
\section{Related Work}
\label{sec:related}
\vspace{-1pt}
\subsection{Adversarial Attacks}
\vspace{-1pt}
Over the past few years, extensive efforts have been made in developing new methods to generate adversarial samples~\cite{chen2017zoo,Ghosh2018Resisting,moosavi-dezfooli2016deepfool,liu2018adv,wong2020fast,zhang2019theoretically,dong2020benchmarking,2020Adversarial}. 
Szegedy \etal~\cite{szegedy2013intriguing} proposed L-BFGS algorithm to craft adversarial samples and showed the transferability property of these samples. 
Goodfellow \etal~\cite{goodfellow2014explaining} proposed Fast Gradient Sign Method (FGSM), a fast approach for generating adversarial samples by adding perturbation proportional to the sign of the cost functions gradient. 
Rather than adding perturbation over the entire image, Papernot \etal~\cite{papernot2016limitations} proposed Jacobian Saliency Map Approach (JSMA), which utilized the adversarial saliency maps to perturb the most sensitive input components.
Kurakin \etal~\cite{kurakin2016adversarial} extended the FGSM algorithm as the Basic Iterative Method (BIM), which recurrently adds smaller adversarial noises.
Madry \etal~\cite{madry2017towards} proposed the attack Projected Gradient Descent (PGD) method by extending the BIM with random start point.
Carlini \etal~\cite{carlini2017towards} proposed an efficient method (\textit{i.e.}, CW attack) to compute good approximations while keeping low computational cost of perturbing examples. It further defined three similar targeted attacks based on different distortion measures ($L_{0}$, $L_{2}$, and $L_{\infty}$). 
It is to be noted that all the above mentioned attacks are white-box attacks that craft adversarial examples based on the input gradient. In the classical black-box attack, the adversarial algorithm has no knowledge of the architectural choices made to design the original architecture. 
There are different ways to generate adversarial samples~\cite{dong2018boosting,ilyas2018black,brendel2018decision,xie2019improving,dong2019evading,cheng2018query,li2019nattack,dong2019efficient} under black-box schemes. Since this paper focuses on the white-box attacks, for more detailed information, readers may refer to~\cite{akhtar2018threat}.

\subsection{Adversarial Defenses}

This line of works focus on developing robust deep models to defend against adversarial attacks~\cite{2020ROSA,Ferrari2022,Tong2021}.
Goodfellow \etal~\cite{goodfellow2014explaining} proposed the first adversarial defence method that uses adversarial training, in which the model is re-trained with both adversarial images and the original clean dataset.
A series of work~\cite{madry2017towards,zhang2019theoretically,kannan2018adversarial} follow this adversarial training, but investigate different adversarial attacks to generate different adversarial data. 
Pappernot \etal~\cite{papernot2017extending} extended defensive distillation~\cite{papernot2016distillation} (which is one of the mechanisms proposed to mitigate adversarial examples), to address its limitation. 
They revisited the defensive distillation approach and used soft labels to train the distilled model. The resultant model was robust to attacks. 
Liang \etal~\cite{liang2017detecting} proposed a method where the perturbation to the input images are regarded as a kind of noise and the noise reduction techniques are used to reduce the adversarial effect.
In their method, classical image processing operations such as scalar quantization and smoothing spatial filters were used to reduce the effect of perturbations. 
Bhagoji \etal~\cite{bhagoji2017dimensionality} proposed dimensionality reduction as a defense against attacks on different machine learning classifiers. 
Another effective defense strategy in practice is to construct an ensemble of individual models~\cite{kurakin2018adversarial}. 
Following this idea, Liu~\etal~\cite{liu2018towards} proposed the random self-ensemble method to defend the attacks by averaging the predictions over random noises injected to the model.
%
%
Pang \etal~\cite{pang2019improving} proposed to promote the diversity among the predictions of different models by introducing an adaptive diversity-promoting regularizer. 
However, these methods do not have an ideal robustness metric to help them correctly evaluate and improve their performance.
\subsection{Robustness Metrics}
With the development of adversarial attacks, there is a need for a robustness metric that quantifies the performance of a DNN against adversarial samples. 
%
%
%
A straightforward method is to use a specific attack method to find the adversarial examples, and use the distortions of adversarial examples (\ie, upper bound of $\Delta_{p,min}$) as the model robustness metric.
For example, Bastani \etal~\cite{bastani2016measuring} proposed a linear programming formulation to find adversarial examples and directly use the $l_p-$distortion as the robustness metric.
Moosavi-Dezfooli \etal~\cite{moosavi-dezfooli2016deepfool} proposed to compute a \textit{minimal} perturbation for a given image in an iterative manner, in order to find the \textit{minimal} adversarial samples across the boundary. They then define all the \textit{minimal} perturbation expectation over the distribution of data as the robustness metric. 
Other methods focus on estimating the lower bound of $\Delta_{p,min}$ and use it as the evaluation metric.
Weng \etal~\cite{Weng2018Towards} exploited the ReLU property to bound the activation function (or the local Lipschitz constant) and provided two efficient algorithms (Fast-Lin and Fast-Lip) for computing a certified lower bound.
Zhang \etal~\cite{NIPS2018_7742} proposed a general framework CROWN for computing a certified lower bound of minimum adversarial distortion and showed that Fast-Lin algorithm is a special case under the CROWN framework.
Recently, a robustness metric called CLEVER~\cite{weng2018evaluating} was developed, which first estimates local Lipschitz constant using extreme value theory and then computes an attack-agnostic robustness score based on first order Lipschitz continuity condition. It can be scaled to deep networks and large datasets. However, the lower bound estimation of CLEVER is often incorrect and is time-consuming. Therefore, it is hard to be a robust and effective adversarial robustness metric.

\section{Methodology}
\label{sec:method}

\begin{figure*}[htb]
	\begin{subfigure}[b]{0.5\linewidth}
		\centering
		\centerline{\includegraphics[width=6cm, height=5cm]{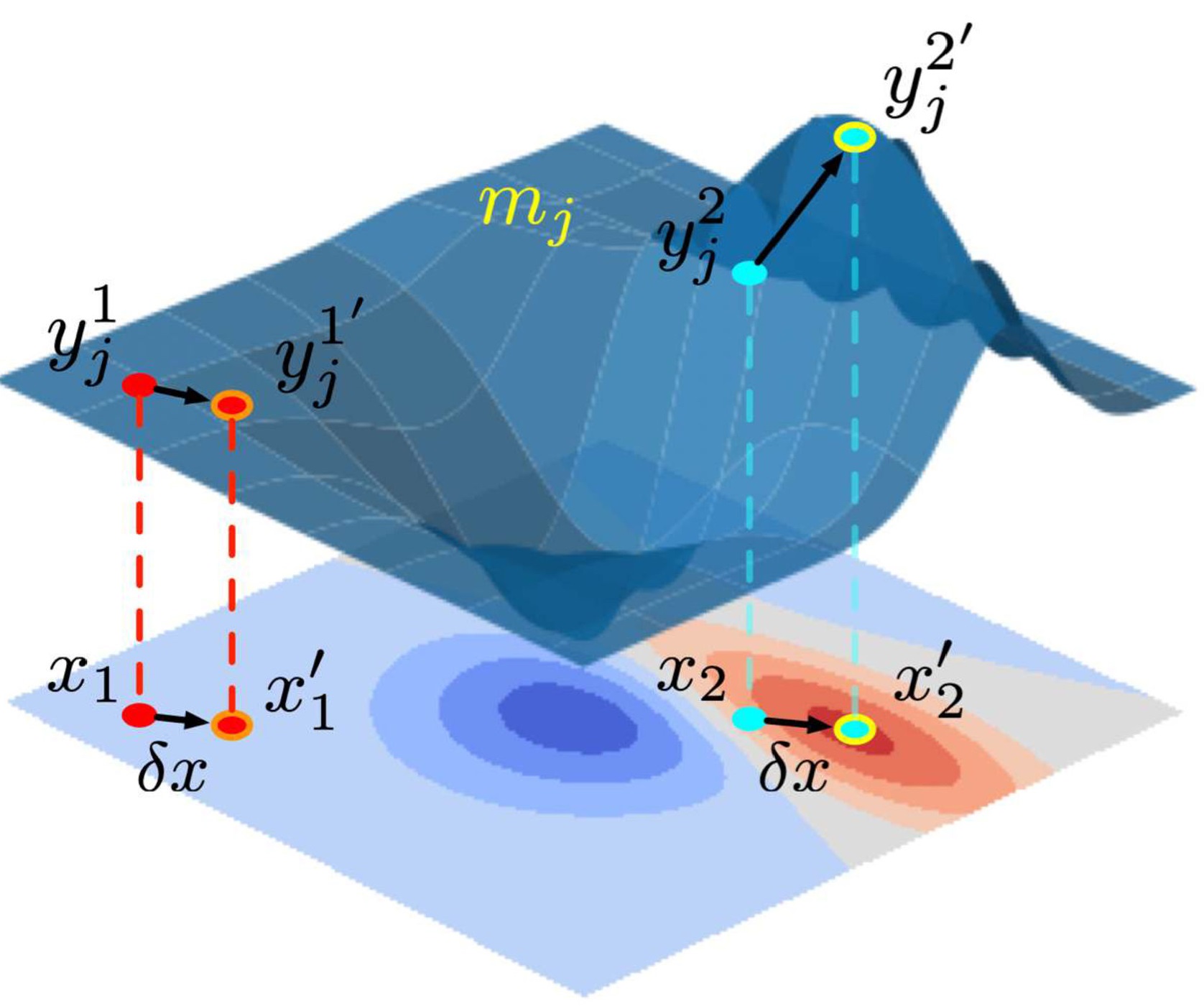}}
		\caption {3D Hypersurface} 
		\label{fig:geo_1}
	\end{subfigure}%
	\hfill
	\begin{subfigure}[b]{0.5\linewidth}
		\centering
		\centerline{\includegraphics[width=6cm, height=5cm]{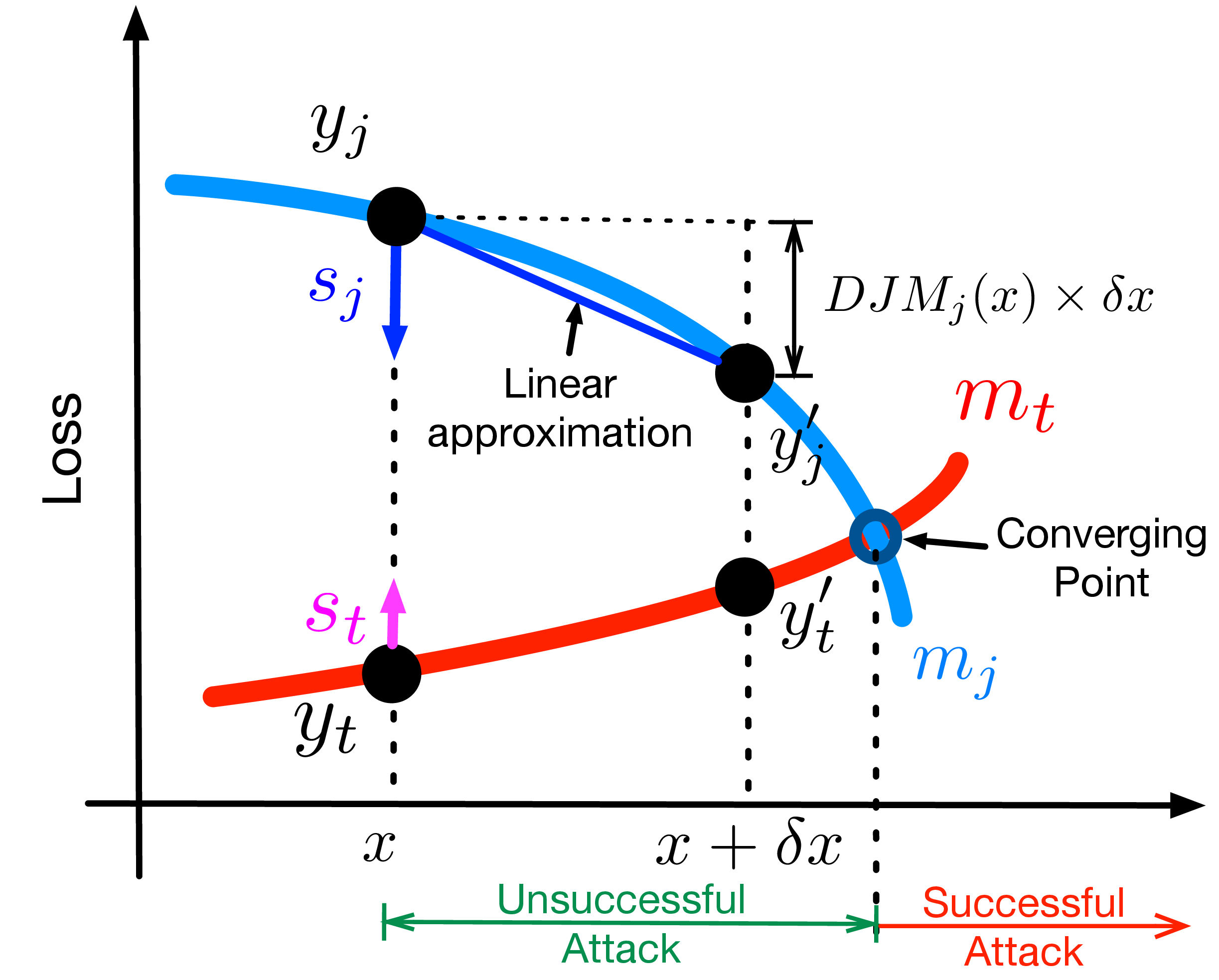}}
		\caption{Intuitions behind ACTS}
		\label{fig:geo_2}
	\end{subfigure}%
	\caption{(a) An example of 3D hypersurface, (b) intuition behind our ACTS.}
	\label{acts—fig}
\end{figure*}

\subsection{Adversarial Converging Time Score}
\label{sec:SRM}
\noindent \textbf{Adversarial Attacks in Image Classification} Given an $M$ dimensional input $x \in \mathbf{R}^M$ and a K-class classification loss function $D: \mathbf{R}^M \rightarrow \mathbf{R}^K$, the predicted class label $t$ of the input $x$ is defined as:  
\begin{equation}
\label{eq:cl}
t = C(x) = \underset{j}{\mathrm{argmin}}\{y_j \mid y_j \in \mathbf{R}^1 \},
\end{equation}
where $y_j$ is the $j$th element of the $K$-dimensional output of $D(x)$. From geometrical point of view, $y_j$ can be regarded as a point on a $M+1$ dimensional hypersurface $m_j$ (See Fig.~\ref{acts—fig} (a)). 

Since DNN-based classifiers are typically non-linear systems, which is true for all state-of-the-art DNN models. In this case, the hypersurfaces $m$ defined by $D$ are also non-linear systems. Thus, local areas around different points on a hypersurface $m_j$ have different curvatures, which results in that different inputs would have different sensitivity to the same added noise $\delta{x}$. As shown in Fig.~\ref{acts—fig} (a), the changes on a hypersurface $m_j$ driven by the same $\delta{x}$ are significantly different in terms of magnitude. Inspired by this insight, we propose a novel Adversarial Converging Time Score (ACTS) as an instance-specific adversarial robustness metric. The key to the proposed ACTS is that the sensitivity is mapped to the ``time'' required to reach the converging curve (\textit{i.e.}, decision boundary) where a clean sample is converted to an adversarial sample. We first introduce the proposed ACTS in detail. Then, we provide a toy-example to validate the proposed approach.

\noindent \textbf{Adversarial Converging Time Score (ACTS)}
To easily convey the intuitive idea of the proposed ACTS, we use 1D input domain and 2D hypersurface (\textit{i.e.}, lines). Fig.~\ref{acts—fig} (b) shows our idea intuitively. As we can see, based on Eq.~\ref{eq:cl}, the original input $x$ is classified as $t$th class since $y_t$ is in a lower position than $y_j$ in the loss domain. Although adding a noise $\delta{x}$ to $x$ results in two new positions $y_t'$ and $y_j'$, the predicted label of $x+\delta{x}$ is not changed (still $t$). If $x+\delta{x}$ passes the converging point, the predicted label of $x+\delta{x}$ changes to $j$. 
From this point of view, the robustness of an input can be reflected by the magnitude of the added $\delta{x}$ for reaching the converging point. 
For a DNN-based classifier, the collection of converging points forms the decision boundary. However, such decision boundary is extremely hard to be estimated, especially in a high-dimensional space. 
Instead, we can look at the converging point from the perspective of loss domain, where the distance between $y_j'$ and $y_t'$ is $0$. 
In other words, the robustness of an input can be reflected by the time used to cover the distance $y_j - y_t$, \ie, the less time it requires, the less robust it is.
%
%
Compared to the decision boundary estimation, estimating the distance $y_j - y_t$ is much easier. 
Hence, we propose the ACTS to estimate such time, which takes the following form:
\begin{align}
\label{eq:s_score}
ACTS& :=\underset{j}{\mathrm{argmin}}(f(\frac{y_j - y_t}{s_t - s_j}))\quad j \in 1\ldots K, j\neq t, \\
f(x) &= \begin{cases} C, & x \leq 0 \\ 
x, & x  > 0  \end{cases} \nonumber 
\end{align}
where $s_j$ and $s_t$ are the moving speeds in the loss domain, which are driven by the added noise $\delta{x}$. 
However, the minus sign in the denominator may be a bit tricky.
An ideal misclassification attack should increase the target error value, results in a positive $s_t$, and it should also decrease the error value of the potential misclassified class, which gives a negative $s_j$ (as shown in Fig.~\ref{fig:geo_2}). 
%
%
Hence, the value of $s_t - s_j$ should always be positive. 
However, the $s_t - s_j$ could be a negative value in the following situations: (a) $s_t$ decreases and $s_j$ increases; 
(b) both of $s_t$ and $s_j$ decrease, but $s_t$ decreases faster; 
(c) both of $s_t$ and $s_j$ increase, but $s_j$ increases faster. 
If any of the above cases happen to an input, it means it is impossible to deliver a successful attack, and hence the ACTS of the specific input is a maximum score $C$. 
The $f(x)$ used in the Eq.~\eqref{eq:s_score} is for this purpose. 
Since ACTS represents the time to cover the distance $y_j - y_t$ with a speed $s_t - s_j$, an input with a smaller ACTS is more vulnerable to an adversarial attack, and vice-versa.

The key to the proposed ACTS is to estimate the moving speed. 
However, a local neighborhood on an output hypersurface is non-linear. It is very challenging to estimate the moving speed directly. To this end, we propose a novel $DJM$ based scheme to estimate the required moving speed, which takes the non-linearity nature of an output hypersurface into account.

\begin{figure*}[htb]
	\begin{subfigure}[b]{0.5\linewidth}
		\centering
		\centerline{\includegraphics[width=6cm, height=4.5cm]{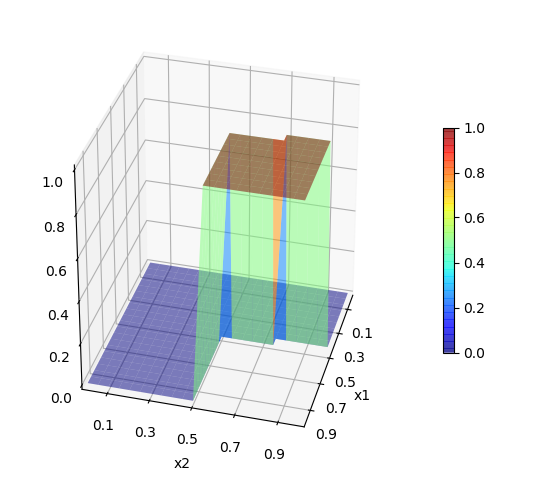}}
		\large{\caption{Model output}}
		\label{fig:toy_output}
	\end{subfigure}%
	\hfill
	\begin{subfigure}[b]{0.5\linewidth}
		\centering
		\centerline{\includegraphics[width=6cm, height=4.5cm]{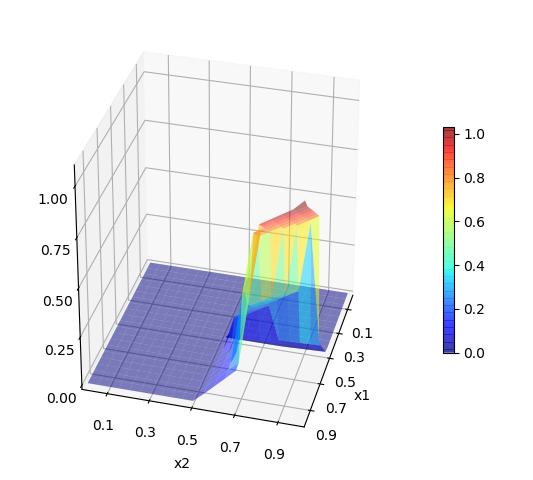}}
		\caption{ACTS distribution}
		\label{fig:toy_nss}
	\end{subfigure}%
	\centering
	\caption{(a) the output of the toy example AND gate model, (b) the ACTS distribution of the toy example AND gate model for all input samples where $x_1 \wedge x_2 = 1$.}
	\label{fig:toy_1}
\end{figure*}

\noindent \textbf{Data Jacobian Matrix} Given an input $x$, the Data Jacobian Matrix (DJM) of $D$ is defined as:
\begin{equation}
\label{eq:DJM}
DJM(x) = \frac{\partial{D(x)}}{\partial{x}} =
\begin{bmatrix}
\frac{\partial{D_j(x)}}{\partial{x_i}}  
\end{bmatrix}_{j\in 1 \ldots K, i\in 1 \ldots M}
\end{equation}
On a hypersurface $m_j$, the $DJM_j(x)$ (i.e., $j$th row of $DJM(x)$) defines the best linear approximation of $D$ for points close to point $x$~\cite{wikiJM}. Therefore, with $DJM(x)$, a small change $\delta{x}$ in the input domain of $D$ can be linearly mapped to the change on the hypersurfaces $m_j$. Mathematically, it can be described as:
\begin{equation}
\label{eq:DJM_LT}
D(x+\delta{x}) = D(x) + DJM(x)\times\delta{x} + \delta{e},
\end{equation}
where $\delta{e} \in R^K$ is the approximation error. Essentially, the $DJM(x)$ is very similar to the gradient backpropagated through a DNN during a training process. The only difference is $DJM(x)$ differentiates with respect to the input $x$ rather than network parameters.

\noindent \textbf{One-step attack}
\label{sssec:cf}
Based on the Eq.~\eqref{eq:DJM_LT}, with an input $x$ and an added noise $\delta{x}$, the original point $y_j (a.k.a., D_j(x))$ is shifted to the point $y_j'$ on the hypersurface $m_j$, and the approximated shifted position of $y_j'$ can be estimated as (shown in Fig.~\ref{acts—fig} (b)):
\begin{equation}
\label{eq:DJM_tan}
y_j' \approx D_j(x) + DJM_j(x)\times\delta{x}, 
\end{equation}
where $DJM_j(x)$ is the $j$th row of the $DJM(x)$. For one-step attack (\textit{e.g.}, FGSM), $\delta{x}$ can be regarded as a vector $\vec{d}$. The direction of $\vec{d}$ is fixed and only the length of $\vec{d}$ varies for delivering a successful attack. Therefore, the moving speed $s_j$ from point $y_j$ to $y_j'$ on the surface $m_j$ driven by the shift $\delta{x}$ in the input domain can be estimated as:
\begin{equation}
\label{eq:DJM_speed}
s_j=\frac{y_j' - y_j}{\|\delta{x}\|} \approx \frac{DJM_j(X)\times\delta{x}}{\|\delta{x}\|}\quad j \in 1\ldots K.
\end{equation}
It is worth to mention that the $DJM$ is an linear approximation for a small $\delta{x}$. The approximation accuracy decreases while $\delta{x}$ increases.

\noindent \textbf{Multi-step attack}
In a multi-step attack (\textit{e.g.}, BIM), each step changes the $\vec{d}$ (\textit{i.e.}, $\delta{x}$) in terms of both direction and length. Compared to one-step attacks, the different directions reveal more curvatures of a local neighborhood, and it increases the probability of discovering a more optimal moving speed to reduce the ``time'' (\textit{i.e.}, added noise) for converting a clean sample to an adversarial one. That is also the reason that multi-step attacks are more effective than one-step attacks. However, the dynamics introduced by multi-step attacks is also troublesome to estimate the desired moving speed. To deal with it, we propose an average moving speed from $y_j$ to $y_j'$ based on all explored directions as follow:
\begin{equation}
\label{eq:DJM_speed_ms}
s_j \approx \frac{1}{N}\sum_{q}^N\frac{DJM_j(x)\times\delta{x_q}}{\|\delta{x_q}\|}\quad j \in 1\ldots K,
\end{equation}
where $N$ is the total steps used in the multi-step attack, and $\delta{x_q}$ is the added noise in the $q$th step. Even though the estimated average speed has limited accuracy, our experiments show the effectiveness of the proposed average speed.

\subsection{Toy Example}
\label{toy_exp}
We design a toy experiment to validate the proposed ACTS, where a simple two-layer feed-forward network was trained to proximate a AND gate. The testing accuracy of the trained model was 99.7\%. Mathematically, we define the AND gate as:
\begin{align*}
x_1 \wedge x_2 &= 1,\quad x_1 \ge 0.5 \text{ and } x_2 \ge 0.5 \\
x_1 \wedge x_2 &= 0,\quad \text{ otherwise}
\end{align*}
\noindent where $x_i \in [0, 1.0]$. Based on this definition, as shown in Fig.~\ref{fig:toy_1} (a), [0.5, 1.0] is the decision boundaries on both $x_1$ and $x_2$ axes, where lower ACTSs are expected. We use FGSM method, with $\epsilon=0.1$, to generate adversarial samples only for the clear sample pairs of $x_1 \wedge x_2 = 1$. For the rest pairs, ACTSs are set to 0. As shown in Fig.~\ref{fig:toy_1} (b), the input pairs closer to the decision boundary have lower ACTSs. Also, we observe an increasing trend in the ACTSs as the input values move further away from the boundary. The maximum ACTS is observed at point $(x_1, x_2) = (1.0, 1.0)$. These observations illustrate the proposed ACTSs is able to reflect the robustness under the FGSM attack. 
\section{Experiments}
\label{sec:exper}
In this section, we first validate the effectiveness and generalization capacity of the proposed ACTS metric against different state-of-the-art DNN models and adversarial attack approaches on the ImageNet~\cite{imagenet_cvpr09} dataset in Section~\ref{sec:vali_results}. 
We then compare the proposed ACTS with CLEVER~\cite{weng2018evaluating} (the only method that can be adapted to deep models and large-scale ImageNet dataset), to show that our method provides a more effective and practical robustness metric in different adversarial settings in Section~\ref{compare with clever}.

\subsection{Experimental Setting}

\noindent{\bf Evaluation dataset and methods}
To evaluate the effectiveness of proposed method on large-scale datasets,
we choose the ImageNet Large Scale Visual Recognition Challenge (ILSVR) 2012 dataset, which has 1.2 million training and 50,000 validation images. 
We evaluate our method on three representative state-of-the-art deep networks with pre-trained models provided by PyTorch~\cite{paszke2017automatic}, \textit{i.e.}, the InceptionV3~\cite{DBLP:journals/corr/SzegedyVISW15}, ResNet50~\cite{he2016deep} and VGG16~\cite{simonyan2014very}, as these deep networks have their own network architectures.
To evaluate the robustness of our method against different attacks, we consider three different state-of-the-art white-box attack approaches, \textit{i.e.}, ~(FGSM~\cite{goodfellow2014explaining}, BIM~\cite{kurakin2016adversarial}, and PGD~\cite{madry2017towards}).

\noindent\textbf{Implementation details} We have implemented our ACTS using the PyTorch framework, and all attack methods using the adversarial robustness PyTorch library: Torchattacks~\cite{paszke2019pytorch}. 
%
A GPU-Server with an Intel E5-2650 v4 2.20GHz CPU (with 32GB RAM) and one NVIDIA Tesla V100 GPU (with 24GB memory) is used in our experiments. For preprocessing, we normalize the data using mean and standard deviation. The images are loaded in the range of $[0, 1]$ and then normalized using a $mean = [0.485, 0.456, 0.406]$ and $ std = [0.229, 0.224, 0.225]$ \cite{paszke2017automatic}.
To control the noise levels in order not to bring any noticeable perceptual differences and show the consistent performance of the proposed ACTS, we add the noise of three different levels: $\epsilon=\{0.00039~(0.1/255)$, $0.00078~(0.2/255)$, $0.00117~(0.3/255)\}$ to the FGSM~\cite{goodfellow2014explaining}, BIM~\cite{kurakin2016adversarial} and PGD~\cite{madry2017towards}, respectively.
We use N1, N2 and N3 to represent these three different noise levels, respectively. We use three steps and set the step size = $\epsilon / 2$ to both BIM~\cite{kurakin2016adversarial} and PGD~\cite{madry2017towards}. We use the untargeted attack setting in all attacks. For each image, we evaluate its top-10 class (\textit{i.e.}, the class with the top-10 maximum probabilities except for the true class, which is usually the easiest target to attack)~\cite{weng2018evaluating} in $DJM$. 

\def\he{35mm}
\begin{figure*}[htb]
	\centering
	\begin{adjustbox}{minipage=\linewidth,scale=1}
		\begin{subfigure}[b]{0.33\textwidth}
			\includegraphics[width=\linewidth, height=\he]{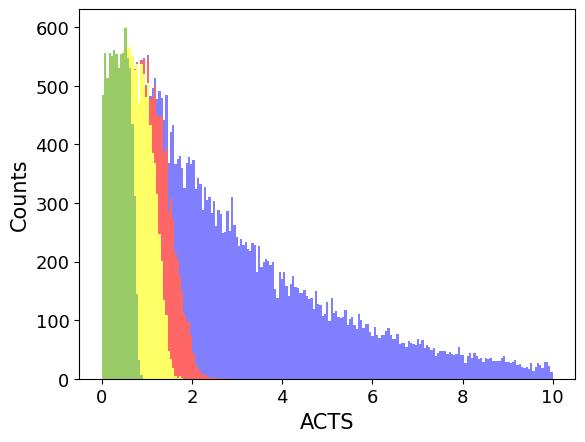}
			\caption{InceptionV3-FGSM}
		\end{subfigure}%
		\begin{subfigure}[b]{0.33\textwidth}
			\includegraphics[width=\linewidth, height=\he]{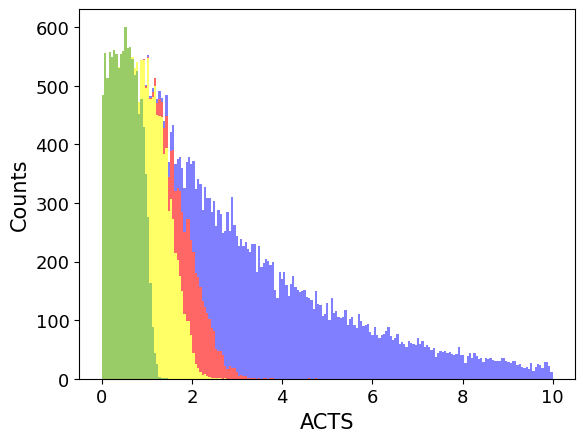}
			\caption{InceptionV3-BIM}
		\end{subfigure}%
		\begin{subfigure}[b]{0.33\textwidth}
			\includegraphics[width=\linewidth, height=\he]{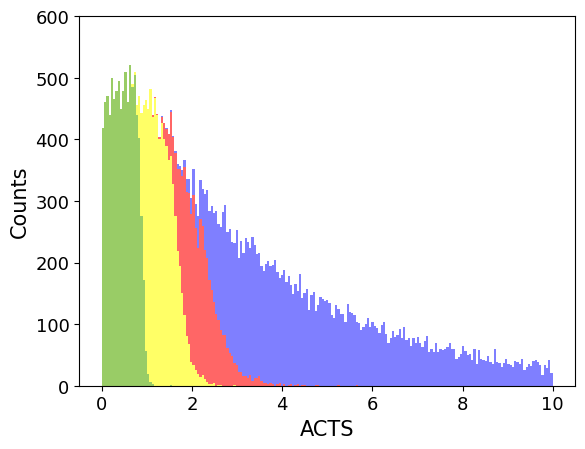}
			\caption{InceptionV3-PGD}
		\end{subfigure}
		\begin{subfigure}[b]{0.33\textwidth}
			\includegraphics[width=\linewidth, height=\he]{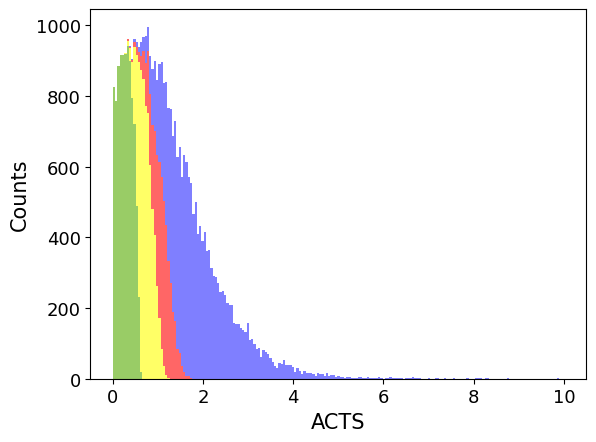}
			\caption{ResNet50-FGSM}
		\end{subfigure}%
		\begin{subfigure}[b]{0.33\textwidth}
			\includegraphics[width=\linewidth, height=\he]{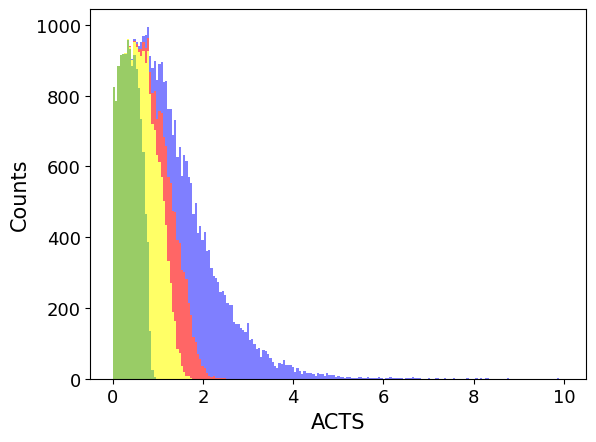}
			\caption{ResNet50-BIM}
		\end{subfigure}%
		\begin{subfigure}[b]{0.33\textwidth}
			\includegraphics[width=\linewidth, height=\he]{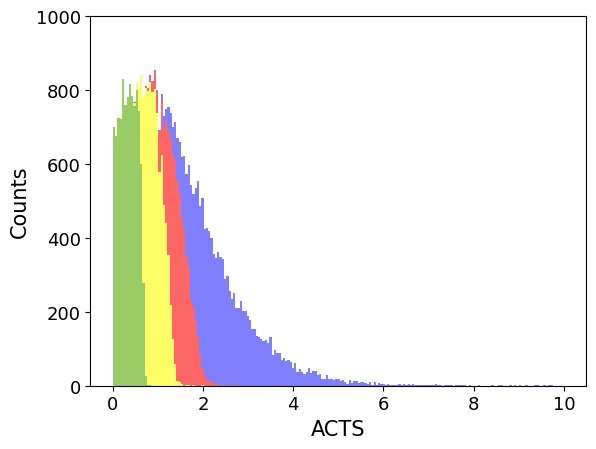}
			\caption{ResNet50-PGD}
		\end{subfigure}
		\begin{subfigure}[b]{0.33\textwidth}
			\includegraphics[width=\linewidth, height=\he]{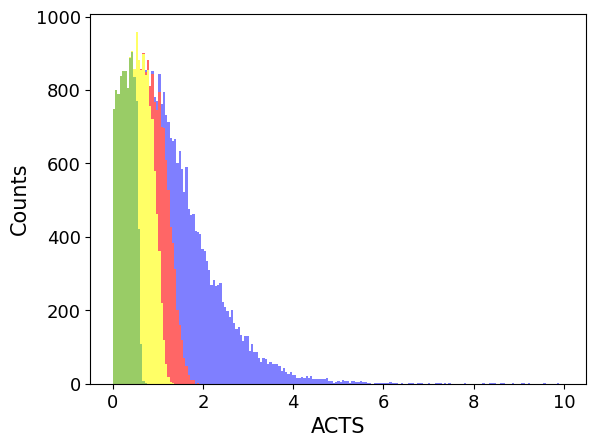}
			\caption{VGG16-FGSM}
		\end{subfigure}%
		\begin{subfigure}[b]{0.33\textwidth}
			\includegraphics[width=\linewidth, height=\he]{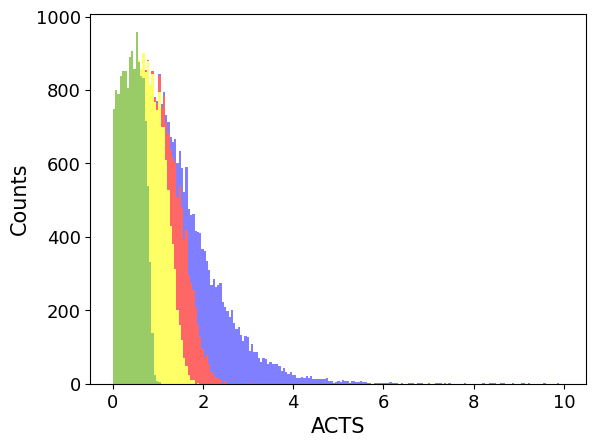}
			\caption{VGG16-BIM}
		\end{subfigure}%
		\begin{subfigure}[b]{0.33\textwidth}
			\includegraphics[width=\linewidth, height=\he]{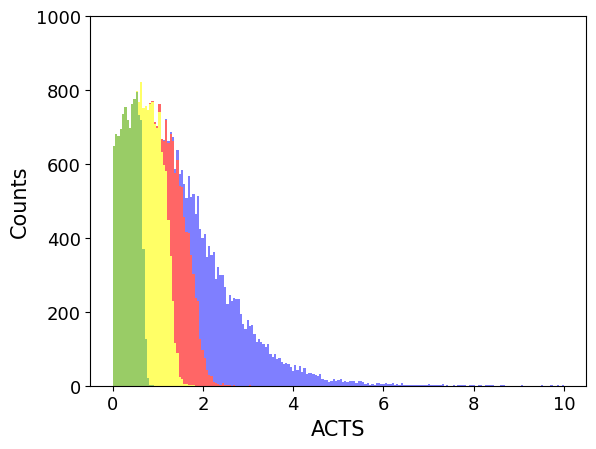}
			\caption{VGG16-PGD}
		\end{subfigure}
		\centering \crule[green!50!white!100]{1em}{0.5em} \footnotesize{(Green):} $\epsilon = \{0.1, 0.00039\}$ 	  				\crule[yellow!50!white!100]{1em}{0.5em} (Yellow):  $\epsilon = \{0.2, 0.00078\}$	
		\crule[red!50!white!100]{1em}{0.5em} (Red):  $\epsilon = \{0.3, 0.00117\}$								\crule[blue!50!white!100]{1em}{0.5em} (Blue): $\epsilon = \{0.05, 0.00020\}$
		\caption{Noise Effectiveness charts for different models under different attacks. Area under the blue color denotes ACTS scores for the correctly classified samples on ImageNet validation dataset . Green, yellow, red colors denoted the ACTS scores of the samples that were successfully attacked. Each color denotes the noise level added to the dataset with respect to the corresponding attack. }
		\label{fig:ATCS}
	\end{adjustbox}
\end{figure*}

\subsection{ACTS Validation Results}
\label{sec:vali_results}
This section evaluates the effectiveness and generalization properties of our proposed ACTS method in various adversarial environments.

\begin{table}[t]
	\centering
	\small
 \renewcommand\arraystretch{0.9}
	\setlength{\tabcolsep}{2mm}{
		\caption{Clean and Adversarial accuracy in different adversarial environments.}
		\vspace{-5pt}
		\begin{tabular}{cccccc}
			\toprule
			\multirow{2}*{Attack}& \multirow{2}*{Model}& \multirow{2}*{Clean}
			&\multicolumn{3}{c}{Adversarial \ Accuracy}\\&&&N1&N2&N3 \\
			\midrule
			\multirow{3}*{FGSM}& InceptionV3& 77.21\%& 61.16\%& 50.03\%&43.05\%\\
			& ResNet50 &76.13\%& 57.45\%& 43.94\%&34.77\%\\& VGG16 &71.59\%& 52.35\%& 37.73\%&27.71\%\\
			\multirow{3}*{BIM}& InceptionV3& 77.21\%& 55.10\%& 43.05\%& 36.68\%\\
			& ResNet50 &76.13\%& 50.08\%& 34.77\%&25.47\%\\& VGG16 &71.59\%& 44.38\%& 27.71\%& 18.46\%\\
			\multirow{3}*{PGD}& InceptionV3& 77.21\%& 60.20\%& 45.86\%& 36.31\%\\
			& ResNet50 &76.13\%& 56.13\%& 39.15\%&26.56\%\\& VGG16 &71.59\%& 51.77\%& 34.48\%& 22.28\%\\
			\bottomrule
		\end{tabular}
		
		\label{Accuracy-Table}}
\end{table}

\noindent{\textbf{Evaluating the effectiveness of ACTS}}
To be an effective adversarial robustness metric, the proposed ACTS should faithfully reflect that the samples with lower ACTS scores are more prone to be attacked successfully than those with higher scores. To validate such property of the ACTS, we design the following experiments. 
First, we apply these three DNN models on the ImageNet validation dataset and selected those correctly classified images. 
Secondly, we estimate the ACTS scores for all the selected images and apply the three chosen attack methods to them. 
It can be seen from Table~\ref{Accuracy-Table} that, {the adversarial accuracy is gradually decreased to a moderate extent with increased levels of noise.}
Third, in order to show the consistent performance of the proposed ACTS, we increase the noise level to N1, N2 and N3 and record the ACTS scores for those who are successfully attacked. 
Fig.~\ref{fig:ATCS} shows the histograms of the three chosen DNN models under different attacks, respectively. 
The blue color indicates the ACTS scores of the images that are correctly classified on ImageNet validation dataset under the initial noise $\epsilon$ = 0.0002, and the other three colors indicate the ones that are attacked successfully with noise level N1, N2 and N3. 
For all the models and attacks, the green, yellow and red regions are always on the very left side of the respective charts. 
This shows the inputs with lower ACTS scores are easier to be attacked successfully.
We can also see that with increased noise levels, images with relatively lower ACTS scores would be attacked successfully first (from green to red).
%
%
In addition, Fig.~\ref{fig:ATCS} and Table~\ref{Accuracy-Table} show that these aforementioned observations are consistent in different adversarial environments (\textit{i.e.}, different DNN architectures and different attacks). Based on the distribution of the obtained
ACTS, we are able to gain a relatively precise intuition about DNNs’ performance under different attack methods. For example, based on the each row figures of Fig.~\ref{fig:ATCS}, it is obvious that ACTS histograms of green, yellow, red colors become much wider range than previous, which indicate BIM and PGD are more powerful attack methods when compared to FGSM attack method. This observations can be confirmed by the corresponding adversarial accuracy rates shown in Table~\ref{Accuracy-Table}.

\def\wdenoising{0.29\linewidth}
\def\hdenoising{0.9in}
\begin{figure*}[htbp]
	\setlength{\tabcolsep}{2.4pt}
	\centering
	\begin{tabular}{ccc}
		\includegraphics[width=\wdenoising, height=\hdenoising]{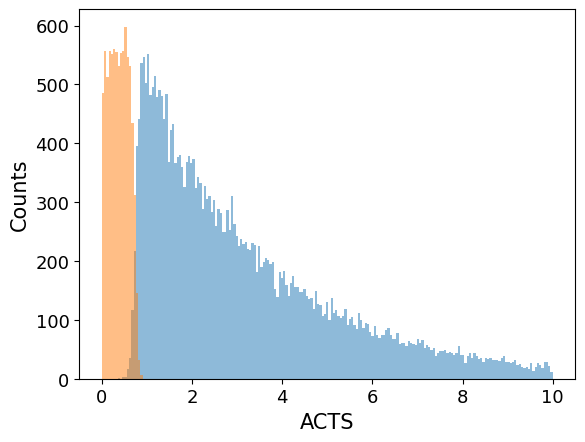}&
		\includegraphics[width=\wdenoising, height=\hdenoising]{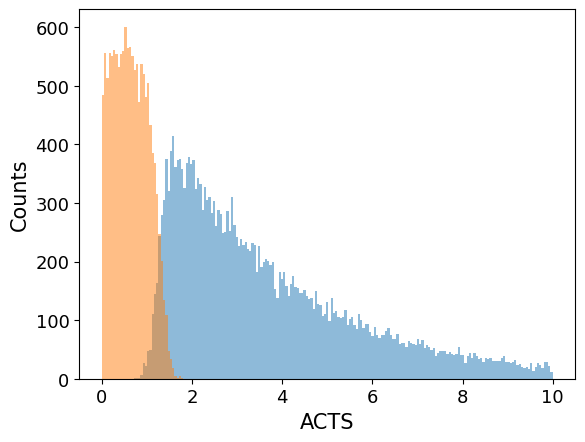}&
		\includegraphics[width=\wdenoising, height=\hdenoising]{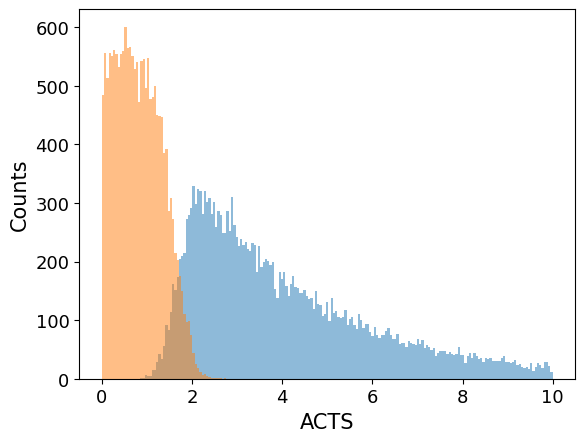}\\
		
		(a) InceptionV3-FGSM-N1 & (b) InceptionV3-FGSM-N2 & (c) InceptionV3-FGSM-N3 \\
		
		\includegraphics[width=\wdenoising, height=\hdenoising]{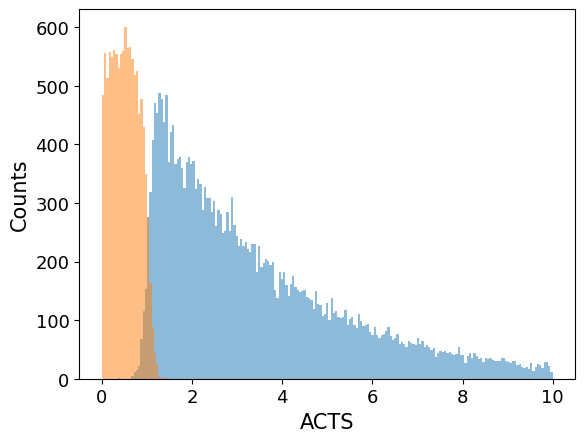}&
		\includegraphics[width=\wdenoising, height=\hdenoising]{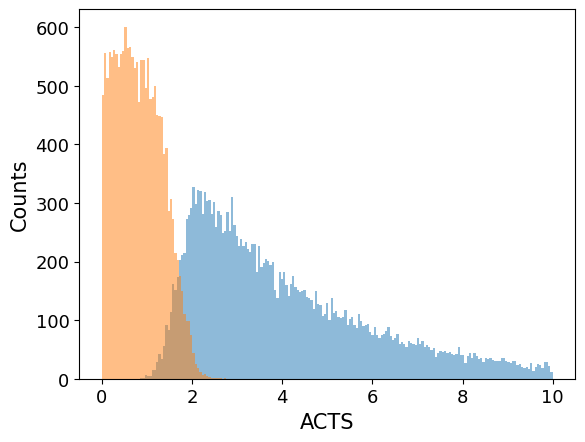}&
		\includegraphics[width=\wdenoising, height=\hdenoising]{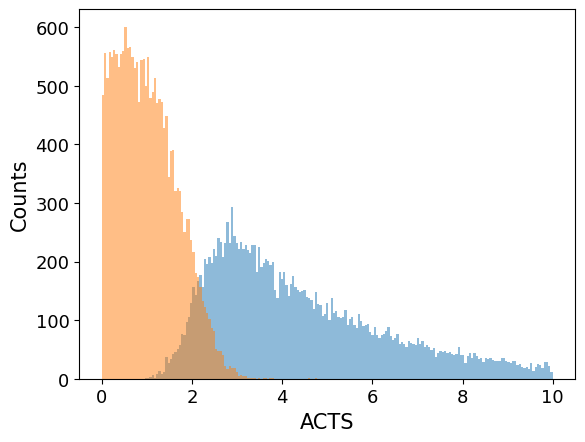}\\
		(d) InceptionV3-BIM-N1 & (e) InceptionV3-BIM-N2 & (f) InceptionV3-BIM-N3 \\
		
		\includegraphics[width=\wdenoising, height=\hdenoising]{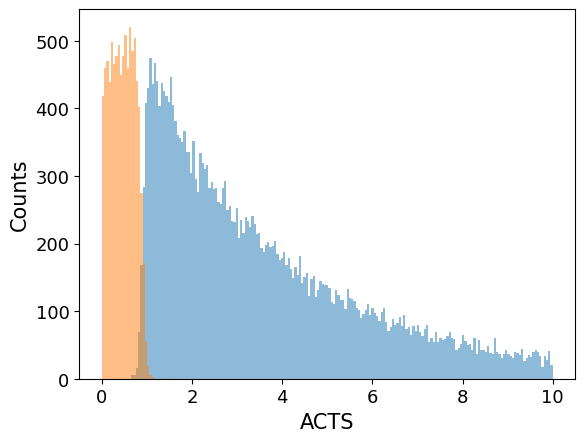}&
		\includegraphics[width=\wdenoising, height=\hdenoising]{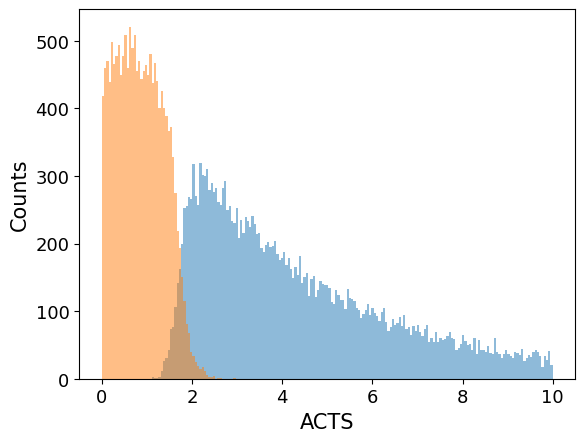}&
		\includegraphics[width=\wdenoising, height=\hdenoising]{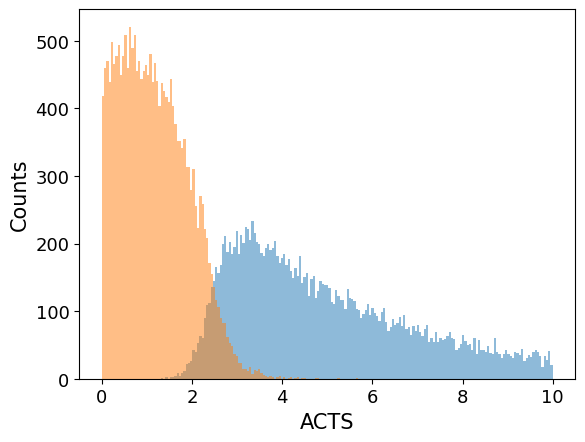}\\
		
		(g) InceptionV3-PGD-N1 & (h) InceptionV3-PGD-N2 & (i) InceptionV3-PGD-N3 \\
		
	\end{tabular}
 \vspace{-5pt}
	\caption{ACTS scores histograms of \textbf{InceptionV3} in different experimental configurations. In each histogram, the orange color indicates the samples that are attacked successfully, and the light blue color indicates the ones that are attacked unsuccessfully. }
	\label{fig:ATCS-overlap-V3}
\end{figure*}

\def\wdenoising{0.29\linewidth}
\def\hdenoising{0.9in}
\begin{figure*}[htbp]
	\setlength{\tabcolsep}{2.4pt}
	\centering
	\begin{tabular}{ccc}
		\includegraphics[width=\wdenoising, height=\hdenoising]{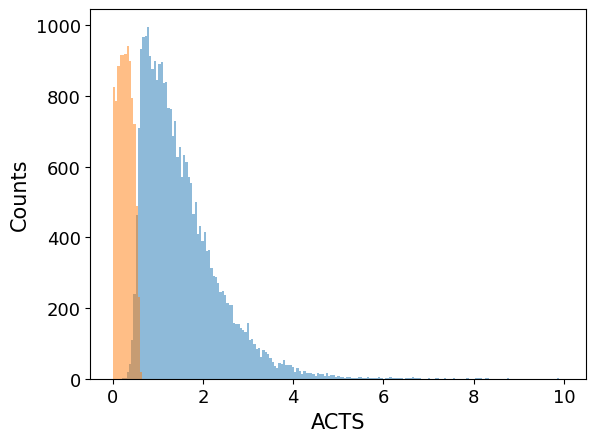}&
		\includegraphics[width=\wdenoising, height=\hdenoising]{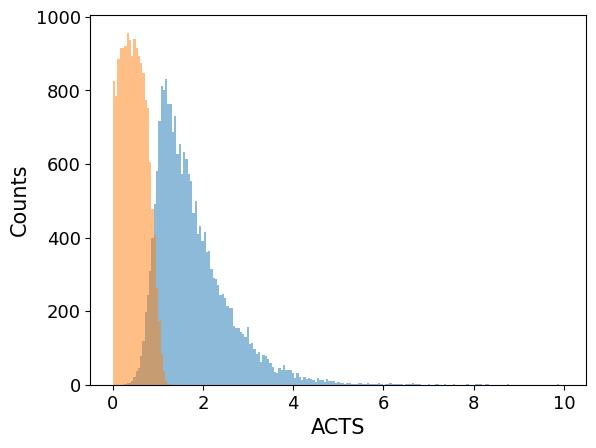}&
		\includegraphics[width=\wdenoising, height=\hdenoising]{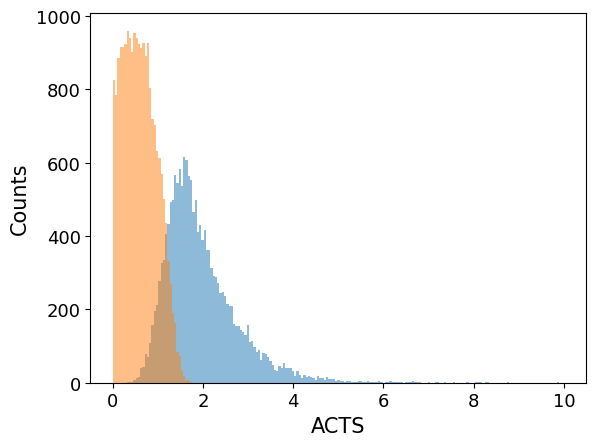}\\
		(a) ResNet50-FGSM-N1 & (b) ResNet50-FGSM-N2 & (c) ResNet50-FGSM-N3 \\
		
		\includegraphics[width=\wdenoising, height=\hdenoising]{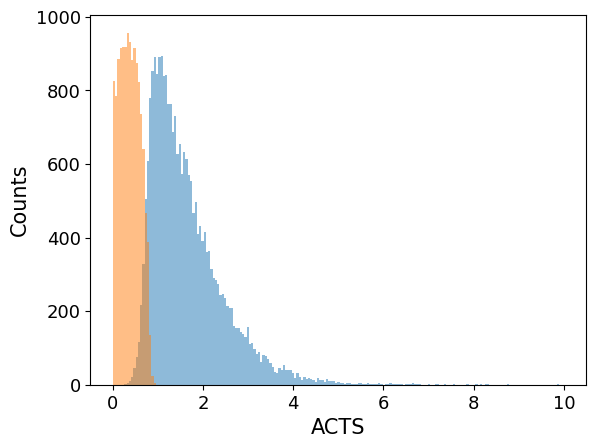}&
		\includegraphics[width=\wdenoising, height=\hdenoising]{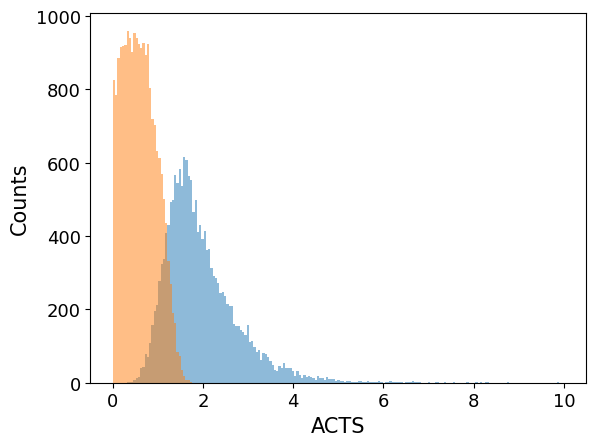}&
		\includegraphics[width=\wdenoising, height=\hdenoising]{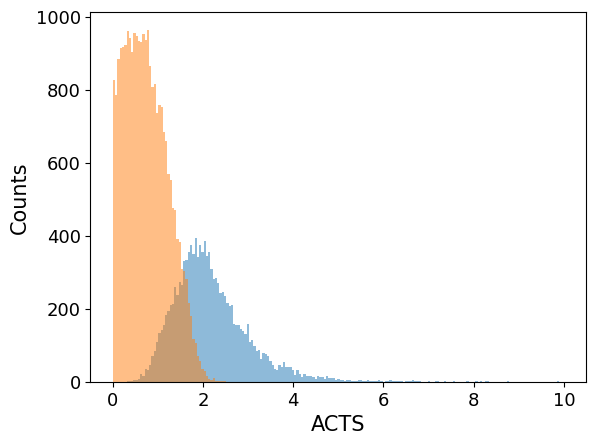}\\
		
		(d) ResNet50-BIM-N1 & (e) ResNet50-BIM-N2 & (f) ResNet50-BIM-N3 \\
		
		\includegraphics[width=\wdenoising, height=\hdenoising]{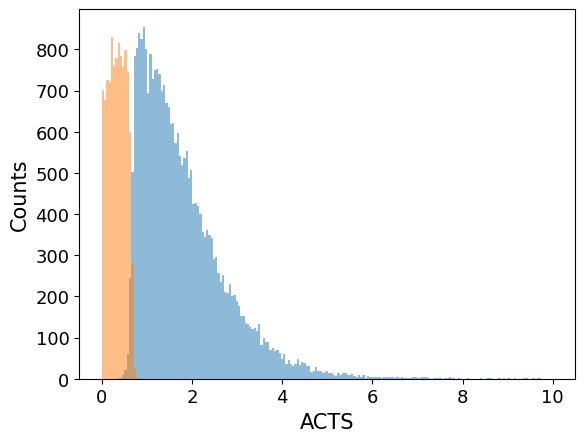}&
		\includegraphics[width=\wdenoising, height=\hdenoising]{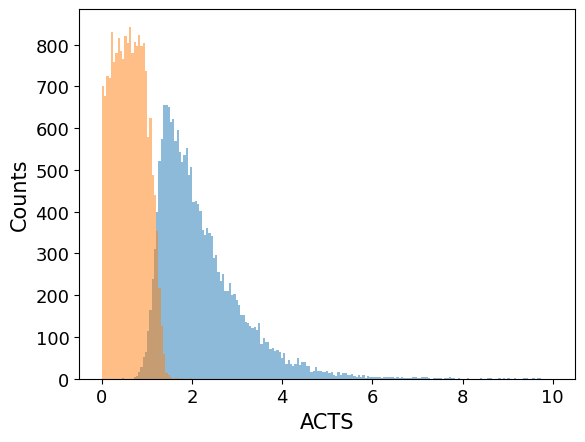}&
		\includegraphics[width=\wdenoising, height=\hdenoising]{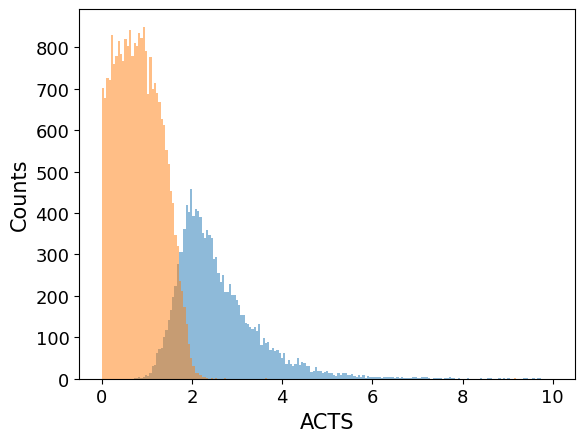}\\
		
		(g) ResNet50-PGD-N1 & (h) ResNet50-PGD-N2 & (i) ResNet50-PGD-N3 \\
		
	\end{tabular}
 \vspace{-5pt}
	\caption{ACTS scores histograms of \textbf{ResNet50} in different experimental configurations. In each histogram, the orange color indicates the samples that are attacked successfully, and the light blue color indicates the ones that are attacked unsuccessfully. }
	
	\label{fig:ATCS-overlap-R50}
\end{figure*}

\def\wdenoising{0.29\linewidth}
\def\hdenoising{0.9in}
\begin{figure*}[htbp]
	\setlength{\tabcolsep}{2.4pt}
	\centering
	\begin{tabular}{ccc}
		\includegraphics[width=\wdenoising, height=\hdenoising]{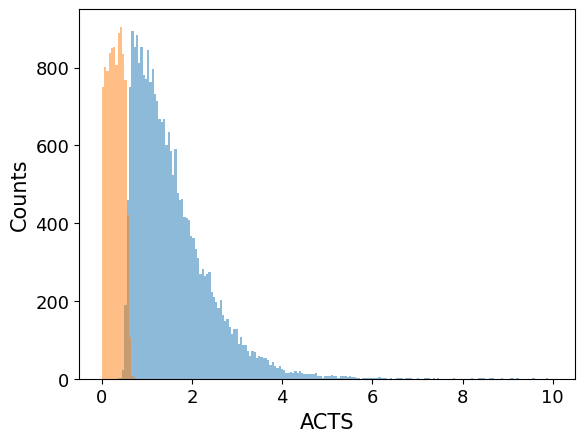}&
		\includegraphics[width=\wdenoising, height=\hdenoising]{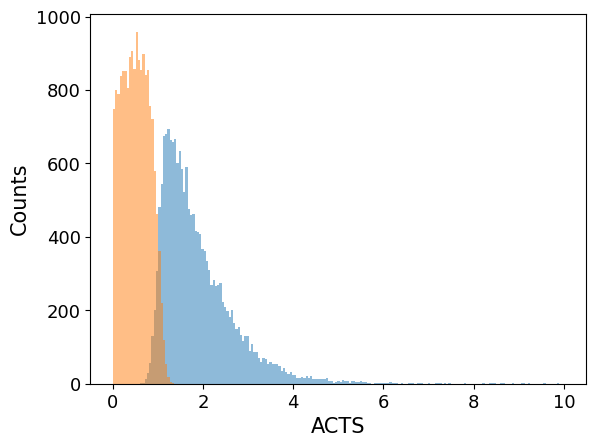}&
		\includegraphics[width=\wdenoising, height=\hdenoising]{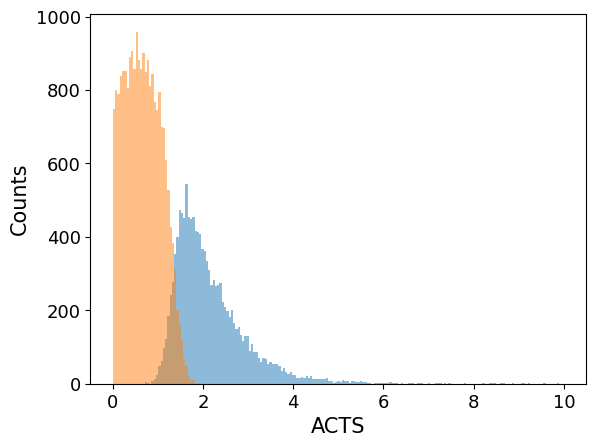}\\
		
		(a) VGG16-FGSM-N1 & (b) VGG16-FGSM-N2 & (c) VGG16-FGSM-N3 \\
		
		\includegraphics[width=\wdenoising, height=\hdenoising]{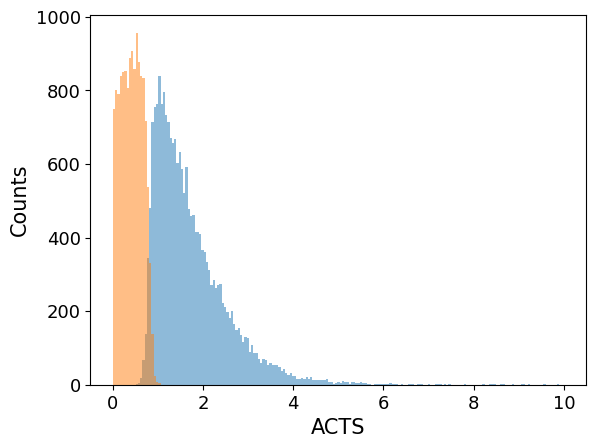}&
		\includegraphics[width=\wdenoising, height=\hdenoising]{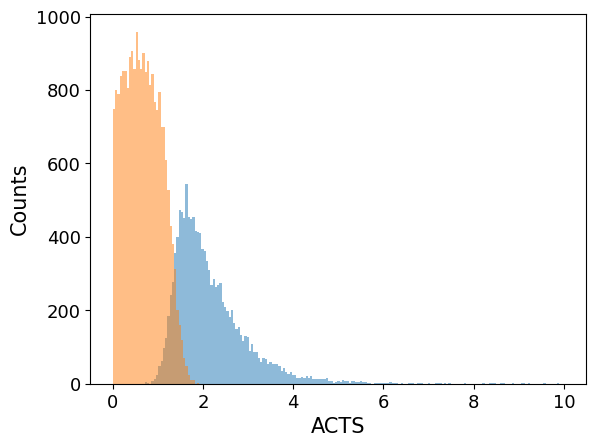}&
		\includegraphics[width=\wdenoising, height=\hdenoising]{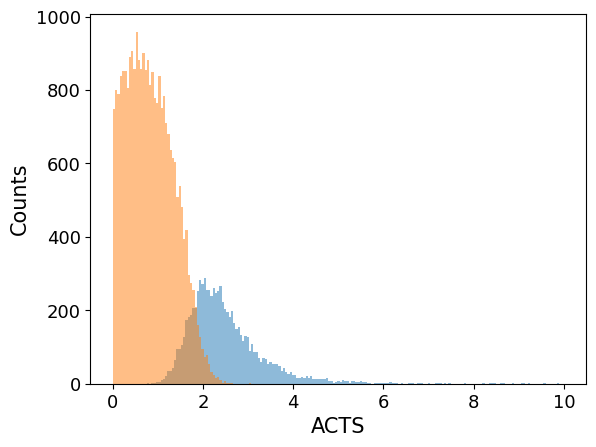}\\
		(d) VGG16-BIM-N1 & (e) VGG16-BIM-N2 & (f) VGG16-BIM-N3 \\
		
		\includegraphics[width=\wdenoising, height=\hdenoising]{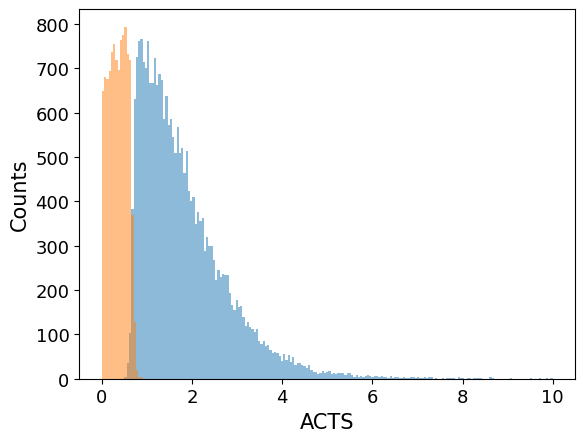}&
		\includegraphics[width=\wdenoising, height=\hdenoising]{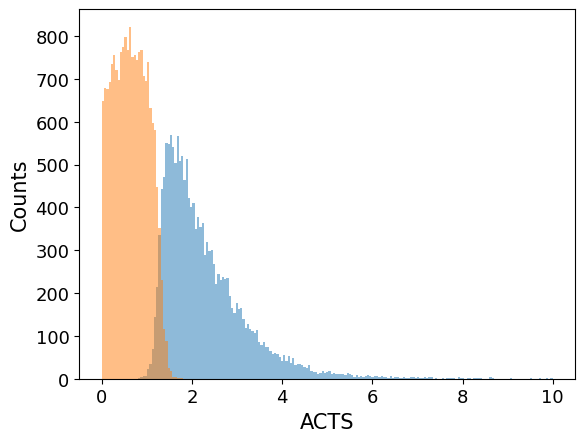}&
		\includegraphics[width=\wdenoising, height=\hdenoising]{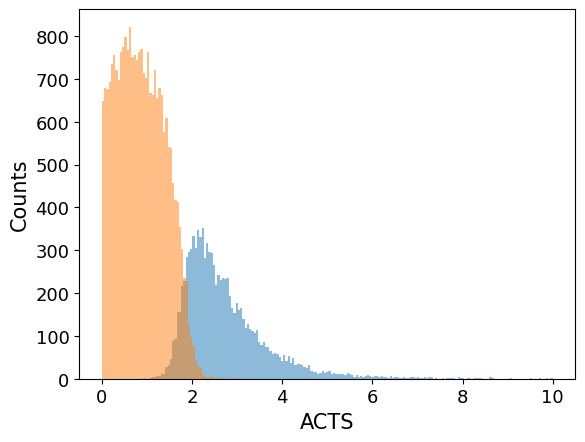}\\
		
		(g) VGG16-PGD-N1 & (h) VGG16-PGD-N2 & (i) VGG16-PGD-N3 \\
		
	\end{tabular}
	\vspace{-5pt}
	\caption{ACTS scores histograms of \textbf{VGG16} in different experimental configurations. In each histogram, the orange color indicates the samples that are attacked successfully, and the light blue color indicates the ones that are attacked unsuccessfully. }
	\label{fig:ATCS-overlap-V16}
	\vspace{-10pt}
\end{figure*}
\begin{table}[hb]
	\centering
	\small
 \renewcommand\arraystretch{0.9}
	\setlength{\tabcolsep}{2mm}{
		\caption{ACTS Overlap\% values in different adversarial environments.}
  \vspace{-5pt}
		\begin{tabular}{lcccc}
			\toprule
			Attack& Model
			&\makecell{Overlap\% \\ N1}
			&\makecell{Overlap\% \\ N2}
			&\makecell{Overlap\% \\ N3}\\
			\midrule
			\multirow{3}*{FGSM}&InceptionV3& 1.46\%& 3.54\%&4.71\%\\
			&ResNet50& 2.95\%& 6.42\%& 9.14\%\\
			&VGG16& 2.14\%&4.29\%&5.71\%\\
			
			\multirow{3}*{BIM}&InceptionV3& 2.53\%& 4.71\%&6.26\%\\
			&ResNet50& 4.89\%& 9.13\%& 10.89\%\\
			&VGG16& 3.02\%&5.71\%&6.47\%\\
			
			\multirow{3}*{PGD}&InceptionV3& 1.33\%& 3.26\%&4.85\%\\
			&ResNet50& 1.72\%& 4.70\%& 6.62\%\\
			&VGG16& 1.87\%& 3.73\%&4.89\%\\
			\bottomrule
		\end{tabular}
		
		\label{tab:overlap-counts-acts}}
\end{table}

\begin{table}[t]
	\centering
	\small
 \renewcommand\arraystretch{0.7}
	\setlength{\tabcolsep}{2.5mm}{
		\caption{Numbers of images in ImageNet for different architectures found in different `attack flip' experimental configurations. `1' means attack flip between different noise level N1 and N2. `2' means attack flip between different noise level N1 and N3. `3' means attack flip between different noise level N2 and N3.}  
  \vspace{-5pt}
		\begin{tabular}{llllllllcc}
			\toprule
			Attack type&
			\multicolumn{3}{c}{FGSM}&\multicolumn{3}{c}{ BIM}&\multicolumn{3}{c}{PGD}\\
			\midrule
			Attack Flip    & 1   & 2 & 3 & 1 & 2 & 3 & 1 & 2 & 3 \\
			\midrule
			InceptionV3  & 0 & 0 &  1 &  1 & 10 & 10&  0  & 0&2   \\
			ResNet50     & 0 & 1  &  1 &  0 &  3 &  5 &  1 &  0 &3  \\
			VGG16     & 0 & 0 &  0 &  0 &  3 & 4 &  2 & 3&8 \\
			\bottomrule
		\end{tabular}
		\label{countflip}}
\end{table} 



In addition to the qualitative results in Fig.~\ref{fig:ATCS}, we also present the quantitative results to show the effectiveness of ACTS. 
Fig.~\ref{fig:ATCS-overlap-V3}, Fig.~\ref{fig:ATCS-overlap-R50} and Fig.~\ref{fig:ATCS-overlap-V16} show the detailed histogram results in different adversarial environments. The orange color indicates the samples that are attacked successfully, and the light blue color indicates the ones that are attacked unsuccessfully. 
Only an ideal robustness metric could separate the two groups without any overlap, and existing approaches may have different overlap regions between the two groups. 
Hence, the size of overlap regions can be leveraged as an indicator to show the effectiveness of a robustness metric. 
For each histogram, we calculate the overlap percentage by $S_o/S_a$, where $S_o$ is the size (\textit{i.e.}, count) of an overlap area, and $S_a$ is the total area of a histogram. 
Hence, for Overlap\%, the lower its values, the better the evaluating results are. All results are shown in Table~\ref{tab:overlap-counts-acts}. As we can see, almost all Overlap\% values are below 10\%. 
In terms of DNN architecture, ACTS shows better performance on InceptionV3 and VGG16. We guess the reason is the local areas on output hypersurfaces of InceptionV3 and VGG16 around the output points of all tested images are flatter (i.e., the radius of curvature is small) than ResNet50. In this case, the $DJM$ provides a more accurate linear approximation.
%
It is worth to mention that the overlap area is getting larger when $\epsilon$ increases in different adversarial environments. It confirms with the limitation of the DJM that the linear approximation accuracy decreases while $\delta{x}$ increases. 
%
%
In the process of statistics, we found an interesting phenomenon called the \textit{attack flip}:
the image with a successful attack at a lower noise level may fail at a higher noise level. 
The result is shown in Table~\ref{countflip}.
\textit{Attack flip} is a good explanation for why there are very small ACTS scores in the overlap at a higher noise level. 
In other words, some small ACTS scores are counted as orange histogram at a lower noise level and then counted as blue histogram at a higher noise level. This flip will result in small ACTS scores in the overlap at a higher noise level.
Besides, another reason is that the limitation of the DJM. The linear approximation accuracy of the DJM decreases while $\delta{x}$ increases which will lead to the error.
\textit{Attack flip} also suggests that the lower bound may not always make sense.

\noindent{\textbf{Evaluating the Generalization of ACTS}}
%
%
In Fig.~\ref{fig:ATCS}, the histogram of each row represent the results of the same model under different attack, and each column represent the results of different models under the same attack method. 
From the results, we can see that ACTS has a good generalization ability across different attack methods and models. 

\noindent{\bf Correlations to CLEVER} We are interested in whether our ACTS align with the CLEVER. To this end, we compute the average score of all the tested images as the CLEVER's reported robustness number.
The higher the CLEVER score, the more robust the model is. 
We also calculate the average ACTS score of all the tested images to represent the robustness of the network. From the results shown in Table~\ref{model_compare}, we get the same ranking correlation to CLEVER. It also demonstrates that models with higher ACTS scores are more robust. The results we obtained are basically consistent with the results in Table (3) (b) of CLEVER (The column of Top-2 Target)~\cite{weng2018evaluating}. Besides, we conclude that VGG16 model with highest scores are more robust than other on test image set. This conclusion can also be found in \cite{su2018robustness}. It is worth to mention that the score distribution may change dramatically on different test image sets.
\begin{table}[htb]
	\centering
	\small
 \renewcommand\arraystretch{0.7}
	\setlength{\tabcolsep}{5mm}{
		\caption{Using CLEVER method and ACTS method to measure the ranking correlation of the robustness of different models.}
  \vspace{-5pt}
		\begin{tabular}{ccc}
			\toprule
			Model&CLEVER&ACTS\\
			\midrule
			VGG16&0.370&4.459\\
			\midrule
			InceptionV3&0.215&3.047\\
			\midrule
			ResNet50&0.126&2.558\\
			\bottomrule
		\end{tabular}
		\label{model_compare}}
\end{table}

\noindent{\bf Determining k} To investigate the impact of the top-\textit{k} class in $DJM$,  for each image, we evaluate its top-\textit{k} class ACTS scores in Table~\ref{tab:overlap-full}. From the results, we can see that with the increase of \textit{k}, the value of Overlap\% changes very slightly. Considering the balance between computational consumption and ACTSs' performance, it is reasonable to set the k to 10. 

\begin{table}[h]
	\centering
	\small
 \renewcommand\arraystretch{0.7}
	\setlength{\tabcolsep}{1mm}{
		\caption{Top-\textit{k} class ACTS Overlap\% values in different environments.}
  \vspace{-5pt}
		\begin{tabular}{lccccc}
			\toprule
			Attack& Model& Metric
			&\makecell{Overlap\% \\ N1}
			&\makecell{Overlap\% \\ N2}
			&\makecell{Overlap\% \\ N3}\\
			\midrule
			\multirow{9}*{FGSM}&\multirow{3}*{ InceptionV3}&ACTS-10&1.56\%&2.08\%&5.2\%\\&& ACTS-20& 1.56\%&2.08\%&5.2\%\\
			&& ACTS-50 & 1.56\%&2.08\%&5.2\%\\
			&\multirow{3}*{ResNet50}&ACTS-10&2.39\%&7.98\%&9.84\%\\&& ACTS-20& 2.39\%& 7.98\%& 9.71\%\\
			&&ACTS-50& 2.39\%& 7.98\%& 9.71\%\\
			&\multirow{3}*{VGG16}&ACTS-10&1.3\%&3.46\%&6.63\%\\&&ACTS-20& 1.3\%&3.46\%&6.77\%\\\
			&&ACTS-50&1.3\%&3.46\%&6.92\%\\
			\midrule
			\multirow{9}*{BIM}&\multirow{3}*{ InceptionV3}&ACTS-10&1.43\%&5.2\%&6.37\%\\&&ACTS-20&1.43\%&5.2\%&6.37\%\\
			&&ACTS-50&1.43\%&5.2\%&6.37\%\\
			&\multirow{3}*{ResNet50}&ACTS-10&4.39\%&9.84\%&10.9\%\\&& ACTS-20&4.26\%&9.71\%&11.17\%\\
			&&ACTS-50&4.26\%&9.71\%&11.17\%\\
			&\multirow{3}*{VGG16}&ACTS-10&2.74\%&3.63\%&5.91\%\\&&ACTS-20& 2.74\%&3.63\%&5.91\%\\
			&&ACTS-50&2.74\%&3.63\%&5.91\%\\
			\midrule
			\multirow{9}*{PGD}&\multirow{3}*{InceptionV3}&ACTS-10&0.91\%&3.25\%&4.81\%\\&&ACTS-20&0.91\%&3.25\%&4.42\%\\
			&&ACTS-50&0.91\%&2.99\%&4.68\%\\
			&\multirow{3}*{ResNet50}&ACTS-10&0.93\%&5.32\%&6.12\%\\&& ACTS-20&0.93\%&5.45\%&6.52\%\\
			&&ACTS-50&0.93\%&5.19\%& 6.38\%\\
			&\multirow{3}*{VGG16}&ACTS-10&1.3\%&3.03\%&4.61\%\\&&ACTS-20&1.44\%&3.17\%&4.61\%\\
			&&ACTS-50&1.44\%&3.17\%& 4.76\%\\
			\bottomrule
		\end{tabular}
		\label{tab:overlap-full}}
\end{table}


\subsection{Comparing With the State-of-the-art CLEVER}
\label{compare with clever}

We compare our method with state-of-the-art method CLEVER in this section. CLEVER score is designed for estimating the lower bound on the minimal distortion required to craft an adversarial sample, and it used $L_2$ and $L_{\infty}$ norms for their validations. 
We follow the setting in \cite{weng2018evaluating} to compute CLEVER $L_2$ and $L_{\infty}$ norms scores for 1,000 images out of the all 5,0000 ImageNet validation set, as CLEVER is more computational expensive. 
The same set of randomly selected 1,000 images from the ImageNet validation set is also used in our method. 
Instead of sampling a high-dimension-space ball, our method only requires normal backpropagations, which is significantly faster than CLEVER. 
Our experiment results in Table~\ref{time-cost} confirm this. 
\begin{table}[h]
	\centering
	\small
	\setlength{\tabcolsep}{3mm}{
		\caption{The average computation time of CLEVER and ACTS on different models for a single image in ImageNet. Blue and red fonts in the third column represent ACTS average computation time under one-step and multi-step attacks, respectively. The fourth column is the corresponding lifting multiple.}
		\begin{tabular}{cccc}
			
			\toprule
			Model&Metric& \makecell{Average\\ Computation \\Time (second)}&\makecell{ACTS\\ speed\_up}\\
			\midrule
			\multirow{2}*{InceptionV3}&CLEVER&331.42&\multirow{2}*{ \textcolor{blue}{6628}/\textcolor{red}{2549}}\\
			&ACTS&\textcolor{blue}{0.05}/\textcolor{red}{0.13}&\\
			\midrule
			\multirow{2}*{ResNet50}&CLEVER&196.25&\multirow{2}*{\textcolor{blue}{4906}/\textcolor{red}{2181}}\\
			&ACTS&\textcolor{blue}{0.04}/\textcolor{red}{0.09}&\\
			\midrule
			\multirow{2}*{VGG16}&CLEVER&286.85&\multirow{2}*{\textcolor{blue}{5737}/\textcolor{red}{2207}}\\
			&ACTS&\textcolor{blue}{0.05}/\textcolor{red}{0.13}&\\
			\bottomrule
			
		\end{tabular}
		\label{time-cost}}
\end{table}

For each image, we calculate its CLEVER and ACTS scores on an NVIDIA Tesla V100 graphics card, the average computation speed of our method is three orders of magnitude faster than CLEVER method on different models. 
We also use the Overlap\% indicator to compare the effectiveness of different robustness metrics, inspired by the ROC curve, which visualizes all possible classification thresholds to quantify the performance of a classifier. 
Since ACTS and CLEVER only care about whether the distribution of image scores are consistent with successful/unsuccessful results in different adversarial environments, 
we can use the Overlap\% indicator as ``mis-classification rate". 
In Table~\ref{tab:overlap-counts_s}, we calculate $L_2$ CLEVER, $L_{\infty}$ CLEVER and ACTS Overlap\% values respectively. 
From the results, we can see that the value range of the $L_2$ CLEVER and $L_{\infty}$ CLEVER Overlap\% is almost in 10\% \~ {} 20\%, and the value range of the ACTS Overlap\% is almost in 0\% \~ {} 10\%. 
CLEVER scores have almost more than twice larger Overlap\% values on average for all testing configurations. 
Even though $L_{\infty}$ CLEVER scores give slightly less Overlap\% values than the ones based on $L_2$ CLEVER scores, ACTS still outperform them in a significant margin with all testing configurations.                                   %
These results indicate that ACTS is a more effective metric than CLEVER in different adversarial environments.

\begin{table}[htb]
	\centering
	\small
    \renewcommand\arraystretch{0.9}
	\setlength{\tabcolsep}{0.5mm}{
		\caption{Comparing ACTS with CLEVER Overlap\% values in different adversarial environments.}
		
		\begin{tabular}{lccccc}
			
			\toprule
			Attack& Model& Metric
			&\makecell{Overlap\% \\ N1}
			&\makecell{Overlap\% \\ N2}
			&\makecell{Overlap\% \\ N3}\\
			\midrule
			\multirow{9}*{FGSM}&\multirow{3}*{ InceptionV3}&$L_2$ CLEVER&14.34\%&15.36\%&17.80\%\\&& $L_{\infty}$ CLEVER& 13.7\%& 15.88\%&17.67\%\\
			&& ACTS &\textbf{ 1.56\%}&\textbf{ 2.08\%}&\textbf{ 5.2\%}\\
			&\multirow{3}*{ResNet50}&$L_2$ CLEVER&17.11\%&18.6\%&19.14\%\\&& $L_{\infty}$ CLEVER& 13.61\%& 15.5\%& 17.25\%\\
			&&ACTS&\textbf{ 2.39\%}&\textbf{7.98\%}&\textbf{9.84\%}\\
			&\multirow{3}*{VGG16}&$L_2$ CLEVER&10.43\%&11.59\%&13.48\%\\&&$L_{\infty}$ CLEVER& 8.99\%&10.43\%&12.75\%\\\
			&&ACTS&\textbf{1.3\%}&\textbf{3.46\%}&\textbf{6.63\%}\\
			\midrule
			\multirow{9}*{BIM}&\multirow{3}*{ InceptionV3}&$L_2$ CLEVER&11.91\%&12.16\%&10.88\%\\&&$L_{\infty}$ CLEVER&11.91\%& 11.65\% & 10.5\%\\
			&&ACTS&\textbf{1.43\%} &\textbf{5.2\%} &\textbf{6.37\%}\\
			&\multirow{3}*{ResNet50}&$L_2$ CLEVER&16.44\%&16.31\%&14.42\%\\&& $L_{\infty}$ CLEVER&13.34\%&15.23\%&13.34\%\\
			&&ACTS&\textbf{4.39\%}&\textbf{9.84\%}&\textbf{10.9\%}\\
			&\multirow{3}*{VGG16}&$L_2$ CLEVER&10.72\%&11.45\%&14.2\%\\&&$L_{\infty}$ CLEVER& 9.13\%&10.58\%&13.19\%\\
			&&ACTS&\textbf{2.74\%}&\textbf{3.63\%}&\textbf{5.91\%}\\
			\midrule
			\multirow{9}*{PGD}&\multirow{3}*{InceptionV3}&$L_2$ CLEVER&12.04\%&11.4\%&11.91\%\\&&$L_{\infty}$ CLEVER&11.01\%&11.27\%&11.78\%\\
			&&ACTS&\textbf{0.91\%}&\textbf{3.25\%}&\textbf{4.81\%}\\
			&\multirow{3}*{ResNet50}&$L_2$ CLEVER&14.15\%&15.77\%&14.82\%\\&& $L_{\infty}$ CLEVER&12.13\%&12.94\%&12.8\%\\
			&&ACTS&\textbf{0.93\%}&\textbf{5.32\%}&\textbf{ 6.12\%}\\
			&\multirow{3}*{VGG16}&$L_2$ CLEVER&7.54\%&10.14\%&13.77\%\\&&$L_{\infty}$ CLEVER&7.68\%&8.7\%&12.17\%\\
			&&ACTS&\textbf{1.3\%}&\textbf{3.03\%}&\textbf{4.61\%}\\
			\bottomrule
		\end{tabular}
		\label{tab:overlap-counts_s}}
\end{table}

\section{Conclusion and Future work}
\label{sec:conclusion}
In this work, we have proposed the Adversarial Converging Time Score (ACTS) as an instance-specific adversarial robustness metric.
ACTS is inspired by the geometrical insight of the output hypersurfaces of a DNN classifier. We perform a comprehensive set of experiments to substantiate the effectiveness and generalization of our proposed metric.
Compared to CLEVER, we prove that ACTS can provide a faster and more effective adversarial robustness prediction for different attacks across various DNN models. 
More importantly, ACTS solves the adversarial robustness problem from a geometrical point of view. We believe it provides a meaningful angle and insight into the adversarial robustness problem, which will help the future work in the same vein.

In the future, we will focus on improving DNN's adversarial performance by leveraging the proposed ACTS. Another interesting direction to look into is extending the ACTS to make it work under black-box attack methods.

\begin{acks}
This work was supported in part by National Key Research and Development Program of China (2022ZD0210500), the National Natural Science Foundation of China under Grant 61972067/U21A20\\491/U1908214, and the Distinguished Young Scholars Funding of Dalian (No. 2022RJ01).
\end{acks}

\bibliographystyle{ACM-Reference-Format}
\bibliography{main}

\end{document}